\documentclass[a4paper]{article}

\usepackage{INTERSPEECH2021}
\usepackage[table,xcdraw]{xcolor}

\title{The Zero Resource Speech Challenge 2021: \\Spoken language modelling}
\name{Ewan Dunbar$^1$, Mathieu Bernard$^2$, Nicolas Hamilakis$^2$, Tu Anh Nguyen$^{2,3}$, Maureen de Seyssel$^2$, Patricia Roz\'e$^2$, Morgane Rivière$^3$, Eugene Kharitonov$^3$, Emmanuel Dupoux$^{2,3}$}

\address{
  $^1$University of Toronto, Canada\\
  $^2$Cognitive Machine Learning (ENS--CNRS--EHESS--INRIA--PSL Research University), France\\
  $^3$Facebook AI Research, France}
\email{ewan.dunbar@utoronto.ca, mathieu.a.bernard@inria.fr, nick.hamilakis562@gmail.com, nguyentuanh208@gmail.com, maureen.deseyssel@gmail.com, patricia.roze@ens.fr, mriviere@fb.com, kharitonov@fb.com, emmanuel.dupoux@gmail.com}

\newcommand{\hs}[1]{\hspace{#1\tabcolsep}}

\begin{document}

\maketitle
\begin{abstract}
 We present the Zero Resource Speech Challenge 2021, which asks participants to learn a language model directly from audio, without any text or labels. The challenge is based on the Libri-light dataset, which provides up to 60k hours of audio from English audio books without any associated text. We provide a pipeline baseline system consisting on an encoder based on contrastive predictive coding (CPC), a quantizer ($k$-means) and a standard language model (BERT or LSTM). The metrics evaluate the learned representations at the acoustic  (ABX discrimination), lexical (spot-the-word), syntactic (acceptability judgment) and semantic levels (similarity judgment). We present an overview of the eight submitted systems from four groups and discuss the main results.
\end{abstract}

\section{Introduction}
Infants are able to learn their native language(s) through observation and interaction before they learn to read and write. They show that it is in principle possible to build a language model in the absence of textual resources, from sensory data only. Being able to reproduce this achievement through automatic means would open up speech and language technology to the majority of the world's languages which do not have enough textual resources to be served by current text-driven approaches.
 
The Zero Resource Speech Challenge Series aims at developing the building block necessary to construct textless AI applications. Previous iterations of the challenge have focused on the discovery of subword units (ZR15,17,19,20) and word units (ZR15,17,20). Here, we wish to push the envelope even further by aiming at learning a language model directly from audio without any annotation nor text. 

Systems are only allowed to use the raw audio of the training set as input; they can use it to discover discrete units from it (pseudo-text) and then train a language model from it, or learn everything end-to-end without discrete units. Following the strategy of the previous challenges, evaluation is done through zero-shot metrics based on human psycholinguistics which do not require any training. Here we are probing four linguistic levels: acoustic, lexical, syntactic and semantic.

We provide baseline systems which are the concatenation of three unsupervised components: self-supervised contrastive representation learning (CPC) \cite{oord2018cpc}, clustering ($k$-means), language modeling (LSTM or BERT). The language models learn on the basis of the pseudo-text derived from clustering the learned representation.
In order to take into account the computing resources of participants, we distinguish submissions by the amount of GPU budget used for training. Accordingly our baseline language models are sorted into a high and a low compute budget. As this benchmark series is about fostering new ideas, not getting the best numbers, we encouraged participants to submit systems in the low budget category.



\section{Methods}

\begin{table*}[]
\def\wexample{6.9cm}
\label{tab:metrics}
\caption{\textbf{Summary description of the four Zero Resouce Benchmark 2021 metrics}. The metrics in light blue use a pseudo-distance $d$ between embeddings ($d_h$ being from human judgments), the metrics in light orange use a pseudo-probability $p$ computed over the entire input sequence.}
\begin{tabular}{p{1.3cm}p{1.6cm}lp{3.6cm}p{\wexample}}
\hline
Linguistic \mbox{level}   & Metrics & Dataset& Task  &Example \\
\hline
\rowcolor[HTML]{E7FAFE} 
acoustic-phonetic  & ABX              &Libri-light& $d(a,x) < d(b,x)?$ \mbox{$a \in A, b\in B,$} \mbox{$x\neq a \in A$} & \parbox[t]{\wexample}{
within-speaker: $(\textrm{apa}_{s_1},\textrm{aba}_{s_1},\textrm{apa}_{s_1})$\\ across-speaker: $(\textrm{apa}_{s_1},\textrm{aba}_{s_1},\textrm{apa}_{s_2})$} \\
\rowcolor[HTML]{FFF7E5} 
lexicon            & spot-the-word    &sWUGGY   & $p(a)\textgreater{}p(b)?$  & \parbox[t]{\wexample}{(brick, $^*$blick)\\(squalled, $^*$squilled)}\\
\rowcolor[HTML]{E7FAFE} 
lexical \mbox{semantics}  & similarity judgment &sSIMI & $d(a,b) \propto d_{h}(a,b) ?$ &\parbox[t]{\wexample}{ (abduct, kidnap) : 8.63\\ (abduct, tap): 0.5 }\\
\rowcolor[HTML]{FFF7E5} 
\rowcolor[HTML]{FFF7E5} 
syntax             & acceptability judgment &sBLIMP& $p(a) > p(b) ?$ & \parbox[t]{\wexample}{(dogs eat meat, $^*$dogs eats meat)\\ (the boy can't help himself, $^*$the boy can't help herself)} \\
\rowcolor[HTML]{E7FAFE} 
\hline
\end{tabular}
\vspace{-1em}
\end{table*}

\subsection{Datasets.}
Participants can use any training set  provided they do not use  textual labels besides the identity of the speaker. We encourage use of the LibriSpeech 960h English dataset \cite{Panayotov2015librispeech}, and for larger models, the clean-6k version of Libri-light  \cite{Kahn_2020_librilight}, a huge collection of speech for unsupervised learning. 

\subsection{Evaluation metrics.}

The evaluation metrics are described in detail in \cite{nguyen2020}, and we only give here high level descriptions for lack of space.

\textbf{Acoustic: the Libri-light ABX metrics.} The ABX metric consists in estimating, for two speech categories $A$ and $B$ (e.g., `bit' and `bet'), the probability that two exemplars $x$ and $a$ of the same category $A$ are closer to one another than two exemplars $x$ and $b$ of different categories $A$ and $B$. 
The score is aggregated across all pairs of triphones like `bit' and `bet', where the change occurs in the middle phoneme. This can be computed within-speaker ($a$, $b$ and $x$ are from the same speaker) or across-speaker ($a$ and $b$ are from the same speaker, and $x$ from a different speaker). To compute this score, participants are required to provide an embedding for each input triphone and to specify a pseudo-distance between acoustic tokens. By default, we provide such a distance, which is the average along a Dynamic Time Warping path realigning $a$, $b$ and $x$ of a distance between embedding frames (angular distance). This metric is agnostic to the dimensionality of the embeddings, can work with discrete or continuous codes, and has been used to compare ASR speech features \cite{schatz2016abx}.  Here, we run it on the pre-existing Libri-light dev and test sets, which has been already used to evaluate several self-supervised models \cite{Kahn_2020_librilight,riviere2020unsupervised}.

\textbf{Lexicon: the sWUGGY spot-the-word metrics.} 
In this task, the models are presented with a pair of spoken tokens: an existing word and a matching nonword.  Participants are to provide a number (probability or pseudo-probability) associated to each acoustic tokens, and models are evaluated on their average accuracy of word-nonword classification based on this probability (chance level: 0.5). The sWUGGY test and development sets consists of 20,000 and 5,000 pairs respectively, with the existing words being part of the LibriSpeech train vocabulary. We also prepared additional OOV-sWUGGY test and development sets consisting of 20,000 and 5,000 pairs respectively, with existing words which do not appear in the LibriSpeech training set. The nonwords are produced with WUGGY \cite{keuleers2010wuggy}, which generates, for a given word, a list of candidate nonwords best matched in phonotactics and syllabic structure, which we additionally filtered for pronouncability using G2P, and for having on average the same unigram and bigram phoneme frequencies as words. Stimuli were produced with the Google Speech API. 

\textbf{Syntax: the sBLIMP acceptability metrics.}  
This part of the benchmark is adapted from BLIMP \cite{warstadt2019blimp}, a set of linguistic minimal sentence pairs of matched grammatical and ungrammatical sentences. Similarly to sWUGGY, the task is to decide which of the  two is grammatical based on the probability of the sentence. The test and dev sets contain 63,000 and 6,300 sentence pairs respectively, with no sentence pair overlap. Stimuli were filtered to contain LibriSpeech vocabulary and for natural prosodic contours, and synthesised as above.

\textbf{Lexical Semantics: the sSIMI similarity metrics.} Here, as in \cite{chung2018speech2vec}, the task is to compute the similarity of the representation of pairs of words and compare it to human similarity judgements. As for the ABX task, participants provide embeddings for input tokens as well as a distance to compute similarity. Here, we provide by default the cosine distance computed over pooled embeddings (with mean, max or min pooling). We used a set of 13 existing semantic similarity and relatedness tests: WordSim-353 \cite{yang2006verb},
WordSim-353-SIM \cite{agirre2009study}, mc-30 \cite{miller1991contextual}, rg-65 \cite{rubenstein1965contextual}, Rare-Word (or rw) \cite{luong2013better}, simLex999 \cite{hill2015simlex},
simverb-3500 \cite{gerz2016simverb}, verb-143 \cite{baker2014unsupervised} , YP-130 \cite{yang2006verb}
and the relatedness-based datasets include MEN \cite{bruni2012distributional}, Wordsim-353-REL \cite{agirre2009study}, mturk-287 \cite{radinsky2011word}, and mturk-771 \cite{halawi2012large}.
All scores were normalised on a 0-10 scale, and pairs within a same dataset containing the same words in different order were averaged.  Pairs containing a word absent from LibriSpeech train set \cite{Panayotov2015librispeech} were discarded. We  selected as development set the mturk-771 dataset and the other 12 datasets were used as test sets, making sure that no pair from the development set was present in any of the test sets.

We then created two subsets of audio files, one synthetic (using the Google API), one natural obtained by retrieving the audio extracts from  LibriSpeech corresponding to each word, following the process presented in \cite{chung2018speech2vec}. In this subset, each word can appear in multiple tokens, providing phonetic diversity; duplicated scores are averaged in the analysis step. The natural subset is smaller than its synthesised counterpart, as we had to discard pairs from the test and dev sets which were not present in the LibriSpeech test and dev sets respectively.  The synthesized subset is composed of 9744 and 705 word pairs for the test and dev sets respectively, and the LibriSpeech subset is composed of 3753 and 309 pairs for the test and dev sets.

\subsection{Toplines and baselines.}

\textbf{The Baseline models}. We build baselines in three steps: acoustic modelling, clustering, and language modelling.

The acoustic model uses Contrastive Predictive Coding (CPC, \cite{oord2018cpc}), where the representation of the audio is learned by predicting the future through an autoregressive model. The CPC model uses a convolutional encoder $g_{\text{enc}}$ to map an input signal $\textbf{x}$ as a sequence of embeddings $\textbf{z}=(z_1,\dots,z_T)$ at a given frame rate. At each time step $t$, a predictor network $g_{\text{pred}}$ takes as input the available embeddings $z_1,\dots,z_t$ and tries to predict the $K$ next future embeddings $\{z_{t+k}\}_{1\leq k\leq K}$ by minimizing a constrastive loss using $\mathcal{N}_t$ negative embedding samples.  We used the PyTorch implementation of CPC\footnote{https://github.com/facebookresearch/CPC\_audio} \cite{riviere2020unsupervised}, which is a modified version of the CPC model with the following architecture:  the encoder $g_{\text{enc}}$ is a 5-layer 1D-convolutional network with kernel sizes of 10,8,4,4,4 and stride sizes of 5,4,2,2,2 respectively, resulting in a downsampling factor of 160, meaning that, for a 16KHz input, the embeddings have a rate of 100Hz.

The predictor $g_{\text{pred}}$ is a multi-layer LSTM network, with the same hidden dimension as the encoder, followed by a 1-layer transformer.  We report results from a 4-layer LSTM, trained on the clean-6k version of Libri-light (see  \cite{nguyen2020} for additional results from a smaller model). Note that we are still able to train low- and higher-budget baselines on the output of this model, as we take the critical and potentially costly part of the pipeline to  be language model training, and therefore calculate the GPU budget based on the language model training time only.


The clustering module uses $k$-means on the outputs of a given hidden layer of the autoregressive model. The clustering is done on the collection of all the output features at every time step of all the audio files in a given training set. After training the $k$-means clustering, each feature is assigned to a cluster, and each audio file can then be discretized to a sequence of discrete units corresponding to the index of the assigned cluster.

We trained the clustering module on the subset of LibriSpeech containing 100 hours of clean speech, using as input the second layer of the CPC model. Our baseline ABX scores are calculated on the framewise $k$-means units, for $k=50$.

The language modeling module takes as input the discretized version of the audio files and is trained with a predictive objective; we used two architectures: LSTM and BERT \cite{devlin2018bert}. We trained a 3-layer LSTM model as a reference small-budget system (22M parameters).
Following \cite{Baevski2020vq-wav2vec}, we trained the BERT model with only the masked token prediction objective. We also followed \cite{Baevski2020vq-wav2vec} by masking a span of tokens in the input sequence instead of a single token (otherwise the prediction would be trivial to the model as discretized units tend to replicate). We masked $M$ consecutive tokens for each span, where $M\sim \mathcal{N}(10,10)$, with a total masking coverage of roughly half of the input tokens (spans may overlap). 
We trained the BERT model using a 12-layer transformer encoder and use this as a reference high-budget system (90M parameters). 
The implementation was done via fairseq (\cite{ott-etal-2019-fairseq}). For further details of the models, see \cite{nguyen2020}.

\textbf{The Topline models. } We trained a BERT model on force-aligned phoneme labels (one per frame) using the gold transcription of the LibriSpeech dataset. We also employed the span masking similarly to the baseline model. 
In addition to the forced-alignment BERT, we also included a BERT model trained on the gold phonetic transcription of the LibriSpeech dataset (no framewise repetitions), with the difference that we only mask one token instead of a span of tokens, since each token is the width of a phoneme rather than a frame. 
For an absolute topline comparison, we used the pretrained RoBERTa large model (\cite{Liu2019RoBERTa}), which was trained on 50K subword units on a huge dataset of total 160GB, 3000 times bigger than the transcription of the LibriSpeech 960h dataset.

\section{Submitted systems and results}

\begin{table*}[ht]
\caption{\textbf{Leaderboard.} Bolded results have the best score in the column among the submitted systems for the given task.}\label{tab:res}
\begin{tabular}{lcc cc c@{\hs{0.4}} cc c@{\hs{0.4}} cc c@{\hs{0.4}} cc}
\hline
                   &       &      &\multicolumn{2}{c}{ABX-with.} &&  \multicolumn{2}{c}{ABX-across}&& sWUGGY & sBLIMP &  & \multicolumn{2}{c}{sSIMI}   \\
\cline{4-5}\cline{7-8}\cline{13-14}
System               & Budget & Set  & clean  & other && clean  & other &&        &           && synth.   & Libri. \\

\hline
\rowcolor[HTML]{E0E0E0} 
Random Baseline      & 0     & dev  & 0.49   & 0.5   && 0.5    & 0.5   && 0.5    & 0.49  && -1.48  & 6.79   \\
\rowcolor[HTML]{E0E0E0} 
                     &       & test & 0.5    & 0.49  && 0.5    & 0.5   && 0.5    & 0.5   && 0.17   & {6.44}   \\
Bert Baseline        & 1536  & dev  & {0.03}   & {0.05}   && {0.04}   & {0.08}  && {0.68}   & {0.56}  && {6.25}   & 4.35   \\
                     &       & test & {0.03}   & {0.05}   && {0.04}   & {0.08}   && {0.68}  & {0.56}  && 5.17   & 2.48   \\
\rowcolor[HTML]{E0E0E0} 
LSTM Baseline        & 60    & dev  & (idem)   & (idem) && (idem) &(idem) && 0.61   & 0.52  && 4.42   & {7.07}   \\
\rowcolor[HTML]{E0E0E0} 
                     &       & test & (idem)   & (idem) && (idem) &(idem) && 0.61   & 0.53  && {7.35}   & 2.38  \\
                     
\hline
                     
BN & 60 &
dev &{0.05 }&{0.09 }&&{0.07 }&{0.13 }&&\textbf{0.64} &\textbf{0.54} &&4.29 &{7.69} \\
&& test  &0.05  &{0.09  }&&{0.07  }&{0.13  }&&\textbf{0.65} &\textbf{0.54}  &&\textbf{9.23  }&-1.14
\\
\rowcolor[HTML]{E0E0E0} 
JC1 & 60 &
dev &\textbf{0.03 }&{0.05 }&&\textbf{0.04 }&{0.08 }&&{0.63} &0.52 &&\textbf{5.90} &\textbf{10.20}  \\
\rowcolor[HTML]{E0E0E0} 
&& test  &\textbf{0.03  }&{0.05  }&&\textbf{0.04  }&{0.08  }&&{0.64}  &0.53  &&2.42  &\textbf{9.02}
\\
JC2 & 60 &
dev &\textbf{0.03 }&\textbf{0.04 }&&\textbf{0.04 }&\textbf{0.07 }&&\textbf{0.64} &{0.53} &&-7.75 &4.60 \\
&& test  &\textbf{0.03  }&\textbf{0.04  }&&\textbf{0.04  }&\textbf{0.07  }&&{0.64}  &0.53  &&{5.15}  &-0.85
\\
\rowcolor[HTML]{E0E0E0} 
AL &  60 &
dev &0.17 &0.20 &&0.25 &0.30 &&0.51 &0.52 &&3.16 &1.79 \\
\rowcolor[HTML]{E0E0E0} 
&& test  &0.17  &0.20  &&0.24  &0.31  &&0.52  &0.52  &&{7.30}&-4.33
\\
TM1 & 60 &
dev &\textbf{0.03 }&{0.05 }&&\textbf{0.04 }&{0.08 }&&0.61 &\textbf{0.54} &&-0.81 &5.45 \\
&& test  &\textbf{0.03  }&{0.05  }&&\textbf{0.04}&{0.09}&&0.61  &\textbf{0.54}  &&{7.00}&-1.47 \\

\rowcolor[HTML]{E0E0E0} 
TM2 & 60 &  dev
&\textbf{0.03 }&{0.05 }&&\textbf{0.04 }&{0.08 }&&0.58 &\textbf{0.54} &&-1.65 &4.81\\
\rowcolor[HTML]{E0E0E0} 
&
&test
&\textbf{0.03 }&{0.05 }&&\textbf{0.04 }&{0.08 }&&0.59 &\textbf{0.54} &&2.89 &-1.67
\\
TM3 &60 & dev
&{0.04 }&{0.06 }&&{0.05 }&{0.10 }&&{0.62} &{0.53} &&-0.17 &{7.07}\\
&
&test
&{0.04 }&{0.06 }&&{0.05 }&{0.11 }&&{0.62} &0.53 &&{5.93 }&0.56
\\

\rowcolor[HTML]{E0E0E0} 
TM4 & 60 & dev &
 {0.04 }& {0.06 }&& {0.05 }& {0.10 }&& 0.60 & {0.53} && -2.10 & {8.89}
\\
\rowcolor[HTML]{E0E0E0} 
& &
test   & {0.04  }& {0.06  }&& {0.05  }& {0.10  }&& 0.60  & 0.53  && {6.74  }& 2.03
\\

\hline
Top: Frame labels & 1536  & dev  & {0}      & {0}    && {0}      & {0}     && {0.92}   & {0.64}  && {7.92}   & 4.54   \\
                     &       & test & {0}      & {0}     && {0}      & {0}     && {0.92}   & {0.63}  && {8.52}   & 2.41   \\
\rowcolor[HTML]{E0E0E0} 
Top: Phone labels      & 1536  & dev  & -      & -     && -      & -     && {0.98}   & {0.67}  && {9.86}   & {16.11}  \\
\rowcolor[HTML]{E0E0E0} \textbf{}\textbf{}
                     &       & test &        &       &&        &       && {0.98}   & {0.67}  && {12.23}  & {20.16}  \\
Top: RoBERTa      & 24576 & dev  & -      & -     && -      & -     && {0.97}   & {0.82}  && {32.28}  & {28.96}  \\
                     &       & test & -      & -     && -      & -     && {0.96}   & {0.82}  && {33.16}  & {27.82}  \\

\hline

\end{tabular}
\end{table*}

Eight systems were submitted. All were low-budget. The systems are in principle less comparable than in previous years, because the choice of training sets was freer, but all systems made use of either the baseline representations or the same choices of corpus for re-training, with the exception of \textbf{HL}, which trained its units on LibriSpeech 100 hrs (clean), a smaller data set.

System \textbf{BN} \cite{niekerk2021} begins with the baseline CPC representations and applies speaker normalization before re-running $k$-means. An LSTM language model architecture is used.

The two systems of \cite{chorowski2021} both use continuous representations for the acoustic evaluation.  \textbf{JC1} uses the baseline CPC representations for the acoustic tests, while \textbf{JC2}  improves on the baseline representations using speaker embeddings, and by pulling the vectors in the direction of a cluster mean. For the other evaluation measures, these submissions use different approaches for each  measure. While this does not directly assess  language modelling, it helps understand the limits of the acoustic representations for finding and encoding different types of linguistic information. Beginning from a ($k$-means) discretized version of the respective acoustic CPC representations, for sWUGGY,  they treat the training corpus as a dictionary and extract a distance to the best match; for sBLIMP, they train an LSTM; and, for sSIMI, they perform segmentation to discover word types. They then train word embeddings using word2vec (\textbf{JC1}) or fast-text (\textbf{JC2}) and use this embedding space to calculate similarity.

System \textbf{HL} \cite{liu2021} trains  representations using Mockingjay, an approach based on bidirectional self-attention \cite{liu2020mockingjay}. These are used as pseudo-labels in a teacher-student training scheme.

Finally,  systems \textbf{TM1}--\textbf{TM4} \cite{maekaku2021} train a first round of CPC representations (using different variants of the baseline CPC models), followed by clustering. These cluster labels are then used as a classification objective for a second trained network. The language models are based on a smaller BERT recipe (28M parameters) we provided \cite{nguyen2020}.

\emph{Acoustic-phonetic.} Results are presented in Table \ref{tab:res}. Since the ABX scores do not rely on the language models, the BERT and LSTM baselines are the same. Most systems improve on the baseline, with the exception of \textbf{HL} (which is the only representation not based on CPC). Multiple systems are tied for first place, but the bigger picture is that, since the baseline is already excellent, the English triphone ABX measure is approaching what one might consider ``solved.''

Note that the ABX task has not changed fundamentally since the first ZeroSpeech challenge, and, importantly, it continues to evaluate the discriminability of allophones only. As the test items are always in the same phonetic context, the test does not measure whether phonemes have a context-invariant representation. For this reason, low ABX scores may not be a sufficient basis for learning language models. We use our other measures to assess this.

\emph{Lexical.}
From  Table \ref{tab:res}, the systems tend to improve quite a bit on the low-budget LSTM baseline (with \textbf{BN} taking the lead). There remains a big gap between the gold transcription, which supports near-perfect performance on identifying real word-forms, and any of the unsupervised representations proposed, giving room for improvement in future challenges.

\emph{Syntactic.} The sBLIMP scores  show smaller improvements over the LSTM baseline than  sWUGGY.  Unlike for sWUGGY, the  toplines show that access to the gold transcription alone is not sufficient: while it helps,  good performance is only achieved by RoBERTa, which notably trains on much more data than our other toplines. We discuss the implications below. Thus, this appears to be a harder metric to do well on.

\emph{Lexical semantics.} sSIMI is  even harder . Even the best-performing reference model, the RoBERTa topline, shows fairly weak correlations with human judgements. And, as with syntax, the gold transcription alone is not sufficient even to reach this weak correlation. More is needed, as evidenced by the large drop in performance for the phone topline compared to RoBERTa. Here, however, it would appear that an additional factor matters:  the temporal resolution of the input representations. The force-aligned topline, which takes frames as input, shows much poorer performance---in some cases worse than the random baseline (note also, for the LibriSpeech test set, the random baseline is actually the best). In spite of the difficulty of the task, the word-discovery approach of \textbf{JC1} appears promising.

\section{Discussion}

In the span of six years and five challenges, astonishing progress has been made on the triphone ABX task. Systems starting from CPC modelling (all but one here) achieve extremely good ABX scores. Previous work has shown that a low ABX score appears to be a very good predictor of good performance on TTS without T \cite{dunbar2019zero,dunbar2020,lakhotia2021generative}. Nevertheless, previous results of the TTS without T task  indicate that finding low-bitrate representations---with coarse-grained resolution  both  temporally and spatially---is a substantial constraint. As hinted at above, future challenges may also introduce more difficult ABX tasks, which require strictly phoneme-level invariance.

The 2021 challenge takes a different route. Rather than adding extrinsic constraints to push representations to be more abstract and text-like, we asked whether unsupervised speech representations could solve one of the  \emph{problems} classically solved by text: language modelling. Results are promising.

Most prominently,  progress has been made in the span of one iteration of this challenge on the sWUGGY metric. This is great news for LM applications that are heavily driven by the lexicon. Thus, the approaches taken here should already be helpful for ASR decoding (note also that some recent self-supervised pretraining approaches to ASR have shown good results without explicit LMs: \cite{baevski2020wav2vec}). There is a big gap between discovered units and text---so, still major room for improvement---but the speed of progress is very encouraging.

Above the word level, things are harder. As for sWUGGY, sBLIMP performance shows promising improvements over the baselines. Still, the gaps with text-based approaches are more pronounced for both sBLIMP and sSIMI. Speech-based language modelling needs to make more progress  before being applicable to tasks like translation and dialogue, for which  good language modelling depends on making  syntactic and semantic predictions.  For syntax, even the topline, character-based models are middling, and the much-improved performance of the pre-trained RoBERTa model comes at a cost: the training set used by that model is the equivalent of many human lifetimes worth of speech. Training on this much speech is not a plausible cognitive model, nor useful  for low-resource tasks. For semantics,  even our best text-based model is far from the target.  

One possible missing piece in both tasks is words. Most  systems  start from representations with the temporal granularity of phonemes or less. RoBERTa may also be limited by its use of word-pieces as input, as even simple word embeddings  obtain  better  correlations than those seen here on metrics   related to sSIMI \cite{dror2017replicability}.  Indeed, \textbf{JC1} demonstrates that explicit term discovery is useful for the sSIMI task. Future challenges may do well to introduce a distinction between fine-grained (``character''-based) and coarse-grained (``word''-based) metrics. 

\section{Summary of contributions}

We have presented the results of eight systems submitted to the Zero Resource Speech Challenge 2021. This serves as a definitive jumping-off point for language modelling from speech, both because of the standard evaluation provided, and because of the novel reference systems submitted. The systems demonstrated such excellent phonetic discriminability scores that the bar must be pushed higher in future challenges in order to be able to demonstrate progress. However, higher-order tasks remain in their infancy, with even the easiest tasks proving  difficult for current unsupervised representations.

\section{Acknowledgements}

The work for MS, PR and for EDupoux and TAN in their EHESS role was supported by the Agence Nationale de la Recherche (ANR-17-EURE-0017 Frontcog, ANR-10-IDEX-0001-02 PSL*, ANR-19-P3IA-0001 PRAIRIE 3IA Institute) and grants from CIFAR (Learning in Minds and Brains) and Facebook AI Research (Research Grant). The work for EDunbar was supported by a Google Faculty Research Award and by the Agence Nationale de la Recherche  (ANR-17-CE28-0009 GEOMPHON,  ANR-18-IDEX-0001 U de Paris, ANR-10-LABX-0083 EFL).

\bibliographystyle{IEEEtran}

\bibliography{mybib}


\end{document}